# A Compared Study Between Some Subspace Based Algorithms

Xing Liu[1], Xiao-Jun Wu[1], Zhen Liu[1], He-Feng Yin[1]

[1] School of Internet of Things Engineering, Jiangnan University, Wuxi 214122, Jiangsu Province, China

**Abstract**: The technology of face recognition has made some progress in recent years. After studying the PCA, 2DPCA, $R_1$-PCA, $L_1$-PCA, KPCA and KECA algorithms, in this paper ECA (2DECA) is proposed by extracting features in PCA (2DPCA) based on Renyi entropy contribution. And then we conduct a study on the 2D$L_1$-PCA and 2D$R_1$-PCA algorithms. On the basis of the experiments, this paper compares the difference of the recognition accuracy and operational efficiency between the above algorithms.

## 1. INTRODUCTION

In the subspace based face recognition algorithms, features are extracted from the training samples. The eigenspace is formed according to the features. Then samples are projected into eigenspace in which the recognition is completed.

Eigenface method for face recognition was proposed by Truk and Pentland [1]. In eigenface method, the image matrices were reshaped into image vectors with a high dimension. In order to improve the efficiency of feature extraction and recognition accuracy, Yang proposed 2DPCA [2]. The $L_2$ norm based PCA algorithm is sensitive to outliers. In [3] and [4], $R_1$-PCA and $L_1$-PCA were proposed by using $R_1$ norm and $L_1$ norm respectively instead of $L_2$ norm. They are not only less sensitive to the outliers but also rotational invariant. Samples were mapped into the high dimensional kernel space in KPCA algorithm [5]. KECA improved KPCA in principal component extraction [6]. Several methods based PCA or CA are proposed. Chen et al. presented a pattern classification method based on PCA and KPCA (kernel principal component analysis), in which within-class auxiliary training samples were used to improve the performance [7]. Liu et al. proposed a 2DECA method, in which features are selected in 2DPCA subspace based on the Renyi entropy contribution instead of cumulative variance contribution [8]. Moreover, some approaches based on linear discriminant analysis (LDA) were explored [9-11].

Inspired by KECA algorithm, we propose ECA and 2DECA algorithms, in which we extract features by selecting eigenvectors contributing most to the Renyi entropy estimate of the data. To solve the problem of calculating on high dimensional image vectors in $R_1$-PCA and $L_1$-PCA, we propose 2D$R_1$-PCA and 2D$L_1$-PCA. Compared to $R_1$-PCA, the most important advantage of 2D$R_1$-PCA is that the convergence rate has been improved significantly. The recognition accuracy of 2D$L_1$-PCA is higher than that of $L_1$-PCA.

This paper is organized as follows: We give a brief introduction to PCA, 2DPCA, $R_1$-PCA, $L_1$-PCA and KECA algorithms in section 2. In section 3, the ECA, 2DECA, 2D$R_1$-PCA and 2D$L_1$-PCA algorithms are proposed. In section 4, the proposed methods are compared through experiments. Conclusions are drawn in section 5.

## 2. RALATED WORKS

### 2.1. PCA

For a given two dimensional image matrix, it is reshaped into an image vector. The training set is formed by these image vectors. Let $X = [x_1, x_2, \cdots, x_i, \cdots, x_n]$ be the training set, whose mean image vector is calculated by the following equation

$$m = \frac{1}{n}\sum_{i=1}^{n} x_i. \qquad (1)$$

The main purpose of PCA is to find orthogonal vectors which can describe features of the training set. These orthogonal vectors can be obtained by the eigenvectors of the covariance matrix, which is defined as

$$C = \frac{1}{n}\sum_{i=1}^{n}(x_i - m)(x_i - m)^T. \qquad (2)$$

The projection matrix is formed by the eigenvectors corresponding to the first $k$ largest eigenvalues of $C$. The training samples are projected into eigenspace by

$$V = W^T X. \qquad (3)$$

In the recognition step, the test samples are projected into the eigenspace first. And then the pattern recognition technology, such as minimum distance classifier, is used to identify the query image.

### 2.2. 2DPCA

In contrast with the PCA method, features are extracted by 2DPCA from the two dimensional image matrices directly. The step of reshaping an image matrix into an image vector is avoided.

Let $F = [F_1, F_2, \cdots, F_i, \cdots, F_n]$ be the training set, where $F_i$ is a matrix with the size of $r \times n'$. The mean matrix of the training set is defined as

$$M = \frac{1}{n}\sum_{i=1}^{n} F_i. \qquad (4)$$

The covariance matrix of the training samples is defined as

$$C = \frac{1}{n}\sum_{i=1}^{n}(F_i - M)(F_i - M)^T. \qquad (5)$$

The projection matrix is formed by the eigenvectors corresponding to the first $k$ largest eigenvalues. The training samples are projected into eigenspace by

$$V = W^T F. \qquad (6)$$

In the recognition step, the test samples are projected into the eigenspace first. And then the pattern recognition technology, such as minimum distance classifier, is used to identify the query image.

### 2.3. R₁-PCA

Let $X = [x_1, x_2, \cdots, x_i, \cdots, x_n]$ be the training set, where $x_i(i = 1, 2, \cdots, n)$ is a $d$ dimensional vector. R₁-PCA algorithm tries to find a subspace by minimizing the error function

$$E_{R_1} = \|X - WV\|_{R_1}, \qquad (7)$$

where $W$ is the projection vector, $V$ is defined as $V = W^T X$, and $\|\cdot\|_{R_1}$ denotes the $R_1$ norm, which is defined as

$$\|X\|_{R_1} = \sum_{i=1}^{n}\left(\sum_{j=1}^{d}x_{ji}^2\right)^{\frac{1}{2}}. \tag{8}$$

The principal eigenvectors of the $R_1$-covariance matrix is the solution to $R_1$-PCA algorithm. The weighted version of $R_1$-covariance matrix is defined as

$$C_r = \sum_i \omega_i x_i x_i^T, \quad \omega_i^{(L_1)} = \frac{1}{\|x_i - WW^T x_i\|}. \tag{9}$$

The weight has many forms of definitions. For the Cauchy robust function, the weight is

$$\omega_i^{(C)} = (1 + \|x_i - WW^T x_i\|^2 / c^2)^{-1}. \tag{10}$$

The basic idea of $R_1$-PCA is starting with an initial guess $W^{(0)}$ and then iterate $W$ with the following equations until convergence

$$\begin{cases} W^{(t+\frac{1}{2})} = C_r(W^{(t)})W^{(t)} \\ W^{(t+1)} = orthoronalize\left(W^{(t+\frac{1}{2})}\right) \end{cases}. \tag{11}$$

The concrete algorithm is given in Algorithm 1.

---

**Algorithm 1:** $R_1$-PCA algorithm
---
**Input:** Data matrix $X$, the subspace dimension $k$.
**Initialize:**
    Compute standard PCA and obtain $W_0$.
    Compute residue $s_i = \sqrt{x_i^T x_i - x_i^T W_0 W_0^T x_i}$.
    Compute $c = median(s_i)$.
Set $W = W_0$.
Update $U$ according to Eq. (11).
Compute $V = W^T X$.
Compute $\Lambda = W^T C_r W$. Check deviation from diagonal.
**Output:** $W, V$.

---

## 2.4. L$_1$-PCA

Let $X = [x_1, x_2, \cdots, x_i, \cdots, x_n]$ be the training set, where $x_i (i = 1,2,\cdots,n)$ is a $d$ dimensional vector. The $L_1$ norm is used in L$_1$-PCA for minimizing the error function

$$E_{L_1} = \|X - WV\|_{L_1}, \tag{12}$$

where $W$ is the projection vector, $V$ is defined as $V = W^T X$, and $\|\cdot\|_{L_1}$ denotes the $L_1$ norm, which is defined as

$$\|X\|_{L_1} = \sum_{i=1}^{d}\sum_{j=1}^{n}|X_{ij}|. \tag{13}$$

In order to obtain a subspace with the property of robust to outliers and invariant to rotations, as a replacement programme, the $L_1$ dispersion using $L_1$ norm in the feature space is maximized by the following equation

$$W^* = \max_{W}\|W^T X\|_{L_1}, subject\ to\ W^T W = I. \tag{14}$$

It is difficult to solve the multidimensional version. Instead of using projection matrix $W$, a column vector $w$ is used in Eq. (14) and the following equation is obtained

$$w^* = \max_w \|w^T X\|_{L_1}, \text{subject to } \|w\|_2 = 1. \tag{15}$$

Then a greedy search method is used for solving (15). The main steps of L$_1$-PCA algorithm are given in Algorithm 2.

---
**Algorithm 2:** L$_1$-PCA algorithm

**Input:** The training set $X = [x_1, x_2, \cdots, x_i, \cdots, x_n]$.
**Initiation:** Initial $w_0$ by random numbers. Then set $w(0) = w(0)/\|w(0)\|_2$, $t = 0$
**Polarity check:** $\forall i \in \{1,2,3,\cdots,n\}$, if $w^T(t)x_i < 0, p_i(t) = -1$, otherwise, $p_i(t) = 1$.
**Flipping and maximization:** Set $t = t + 1, w(t) = \sum_{i=1}^n p_i(t-1)x_i, w(t) = w(t)/\|w(t)\|_2$.
**Convergence check:**
  a) If $w(t) \neq w(t-1)$, go to step 2.
  b) Else if $i$ exists such that $w^T(t)x_i = 0$, set $w(t) = (w(t)+\Delta w)/\|w(t)+\Delta w\|_2$, where $\Delta w$ is a small nonzero random vector. Go to step 2.
  c) Otherwise, set $w^* = w(t)$ and stop.
**Output:** The projection vector $w$.

---

One best feature is extracted by the above algorithm. In order to obtain a $k$ dimensional projection matrix instead of a vector, an algorithm based on the greedy search method is given as follows.

---
**Algorithm 3:** Algorithm for finding a m dimensional projection matrix

$w_0 = \mathbf{0}, \{x_i^0 = x_i\}_{i=1}^n$.

For $j = 1$ to $k$

$\forall i \in \{1,2,3\cdots,n\}, x_i^j = x_i^{j-1} - w_{j-1}\left(w_{j-1}^T x_i^{j-1}\right)$.

Apply the L$_1$-PCA procedure to $X^j = [x_1^j, \cdots, x_n^j]$ to find $w_j$.

End

---

### 2.5. KECA

Let $X = [x_1, x_2, \cdots, x_i, \cdots x_n]$ be the training set, where $x_i, i = 1,2,\cdots,n$ is a $d$ dimensional vector. The polynomial kernel of degree $p$ is used to project samples into the kernel space in this paper, which is defined as

$$k^{'}(x_i, x_j) = (x_i \cdot x_j)^p. \tag{16}$$

The $n \times n$ kernel matrix $K$ is defined as

$$K_{ij} = k^{'}(x_i, x_j), i,j = 1,2,3,\cdots,n. \tag{17}$$

In KECA algorithm, the eigenvectors of $K$ is selected according to the Renyi quadratic entropy, which is given by

$$H(p) = -\log \int p^2(x)\,dx. \tag{18}$$

The precise value of $H(p)$ should not be calculated. It is being used to compare the size of the entropy. So calculate the following equation instead

$$sp = \int p^2(x)\,dx. \tag{19}$$

In order to estimate $sp$, a Parzen window density estimator is given as

$$\hat{p}(x) = \frac{1}{n}\sum_{x_t} k^{'}(x,x_t). \qquad (20)$$

In this way, the calculation of the entropy is associated with the kernel matrix. The eigenvalues of $K$ are $\lambda = [\lambda_1,\cdots,\lambda_n]$, the corresponding eigenvectors are $E = [e_1,\cdots,e_n]$ and $l$ is an $n \times 1$ vector whose elements are equal to one. The approximate value of $sp$ could be calculated by the following equation

$$\hat{sp} = \frac{1}{n^2}\sum_{i=1}^{n} (\sqrt{\lambda_i}e_i^T l)^2. \qquad (21)$$

The eigenvectors corresponding to the first $k$ largest entropy are selected as the features.

## 3. PROPOSED METHODS

### 3.1. ECA

Inspired by the KECA algorithm, we extract features by PCA according to the entropy contribution and the ECA algorithm is proposed. The ECA algorithm is given as follows.

**Algorithm 4:** The ECA algorithm

**Input:** The training dataset $X = [x_1,x_2,\cdots,x_i,\cdots x_n]$.

**Calculate the mean image vector:** $m = \frac{1}{n}\sum_{i=1}^{n} x_i$.

**Calculate the covariance matrix:** $C = \frac{1}{n}\sum_{i=1}^{n} (x_i - m)(x_i - m)^T$.

**Calculate the eigenvalues and eigenvectors:** Calculate the eigenvalues and eigenvectors of $C$ marked as $\lambda = [\lambda_1,\cdots,\lambda_n]$ and $E = [e_1,\cdots,e_n]$ respectively.

**Calculate estimated value of the entropy:** $\hat{V} = \frac{1}{n^2}\sum_{i=1}^{n} (\sqrt{\lambda_i}e_i^T l)^2$.

**Features extraction:** Select eigenvectors corresponding to the entropy contribution as the projection matrix $W$.
**Projection:** $V = W^T X$.
**Output:** $W$ and $V$.

### 3.2. 2DECA

Much similar to ECA, we propose 2DECA by extracting features using 2DPCA according to the entropy contribution. The difference is that the image matrices are used in 2DECA instead of image vectors used in ECA. The 2DECA algorithm is given as follows.

**Algorithm 5:** The 2DECA algorithm

**Input:** The training dataset $F = [F_1,F_2,\cdots,F_i,\cdots F_n]$.

**Calculate the mean image matrix:** $M = \frac{1}{n}\sum_{i=1}^{n} F_i$.

**Calculate the covariance matrix:** $C = \frac{1}{n}\sum_{i=1}^{n} (F_i - M)(F_i - M)^T$.

**Calculate the eigenvalues and eigenvectors:** Calculate the eigenvalues and eigenvectors of $C$

marked as $\lambda = [\lambda_1,\cdots,\lambda_n]$ and $E = [e_1,\cdots,e_n]$ respectively.

**Calculate estimate value of the entropy:** $\widehat{V} = \frac{1}{n^2}\sum_{i=1}^{n}(\sqrt{\lambda_i}e_i^T l)^2$.

**Features extraction:** Select eigenvectors corresponding to the entropy contribution as the projection matrix $W$.
**Projection:** $V = W^T X$.
**Output:** $W$ and $V$.

### 3.3. 2DR₁-PCA

In this paper we propose 2DR₁-PCA algorithm, in which we iterate the projection matrix $W$ with a start matrix $W^{(0)}$ until convergence. Let $F = [F_1, F_2, \cdots, F_i, \cdots F_n]$ be the training set. The R₁ covariance matrix is defined as

$$C_r = \sum_{i=1}^{n} \omega_i F_i F_i^T. \qquad (22)$$

The Cauchy weight is defined as

$$\begin{cases} \omega_i^{(C)} = \left(1 + \|F_i - WW^T F_i\|_F^2 / c^2\right)^{-1} \\ c = median(s_i) \end{cases}. \qquad (23)$$

The residue $s_i$ is defined as

$$s_i = \|F_i - WW^T F_i\|_F. \qquad (24)$$

After obtaining the eigenvectors of $C_r$, the iterative formula is similar to which used in the R₁-PCA algorithm

$$\begin{cases} W^{(t+\frac{1}{2})} = C_r(W^{(t)})W^{(t)} \\ W^{(t+1)} = orthoronalize\left(W^{(t+\frac{1}{2})}\right) \end{cases}. \qquad (25)$$

The 2DR₁-PCA algorithm is given as follows

**Algorithm 6:** The 2DR₁-PCA algorithm
**Input:** The training dataset $F = [F_1, F_2, \cdots, F_i, \cdots F_n]$.
Compute the standard PCA to obtain $W^{(0)}$.
Calculate the covariance matrix $C_r$ according to Eqs. (23) and (24).
Iterate until convergence according to Eq. (25).
**Projection:** $V = W^T X$.
**Output:** $W$ and $V$.

### 3.4. 2DL₁-PCA

The main purpose of L₁-PCA algorithm is finding a vector to maximize Eq. (15). For the two dimensional case, let $F = [F_1, F_2, \cdots, F_i, \cdots, F_n]$ be the training set, where $F_i$ is a $r \times n'$ image matrix. $F$ can be rewritten as $F = [f_1, f_2, \cdots, f_i, \cdots, f_{n \times n'}]$, where $f_i$ is an $r$ dimensional vector.

$$w^* = \max_w \|w^T F\|_{L_1} = \max_w \sum_{i=1}^{n} \|w^T F_i\|_{L_1} = \max_w \sum_{i=1}^{n} \sum_{j=1}^{n'} |w^T (F_i)_j| =$$

$$\max_w \sum_{i=1}^{n \times n'} |w^T f_i|, subject\ to\ \|w\|_2 = 1, \quad (26)$$

where $(F_i)_j$ denotes the $j$th column of the image matrix $F_i$. From Eq. (26) we know that the 2DL₁-PCA can be proposed by using the dataset $F = [F_1, F_2, \cdots, F_i, \cdots, F_n] = [f_1, f_2, \cdots, f_i, \cdots, f_{n \times n'}]$ as the input of algorithms given in section 2.4.

## 4. EXPERIMENTS

The above methods are tested on three databases: the ORL [12], YALE []13 and XM2VTS [14] databases. The recognition accuracy and running time of extracting features are recorded.

The ORL database contains face images from 40 different people and each person has 10 images, the resolution of which is $92 \times 112$. Variation of expression (smile or not) and face details (wear a glass or not) are contained in the ORL database images. In the following experiments, 5 images are selected as the training samples and the rest are selected as the test samples.

The YALE database is provided by YALE University. The YALE database contains face images from 15 different people and each one has 11 images. The resolution of YALE database images is $160 \times 121$. In the following experiments, 6 images are selected as the training samples and the remaining are selected as the test samples.

The XM2VTS database offers synchronized video and speech data as well as image sequences allowing multiple view of the face. It contains frontal face images taken of 295 subjects at one month intervals taken over a period of few months. In the following experiments, 4 images are selected as the training samples and the rest are selected as the test samples.

### 4.1. PCA and 2DPCA

The first group experiments are comparative test of PCA and 2DPCA. The experiments result on the ORL database is shown in Table 1.

Table 1. Experiments result of PCA and 2DPCA on the ORL database.

| Algorithms | Recognition accuracy | Running time |
|---|---|---|
| PCA | 0.9 | 1.375 |
| 2DPCA | 0.91 | 0.2913 |

The running time of 2DPCA is shorter. In this experiment, the training set of PCA is a $10304 \times 200$ matrix, whose covariance matrix is a $10304 \times 10304$ singular matrix. It is difficult to calculate the eigenvalues and eigenvectors.

The size of covariance matrix of 2DPCA is much smaller than that of PCA. It is a $112 \times 112$ matrix according to Eq. (5). The computational complexity of calculating eigenvalues and eigenvectors in 2DPCA is lower than that of PCA. As a result, the efficiency of 2DPCA is much higher than that of PCA.

The PCA and 2DPCA algorithms are all based on statistical laws. The more information contained in the input data, the higher recognition accuracy could be obtained. All information are contained in the image matrix. But some information is lost when image matrices are transformed into image vectors. So the recognition accuracy of 2DPCA is higher than that of PCA.

According to the above analysis, compared with PCA, a higher recognition accuracy and shorter running time would be obtained in 2DPCA algorithm. The experiments result is shown in Table 2, which verifies the above analysis.

Table 2. Experiments result on the YALE and XM2VTS databases.

| Algorithms | YALE | | XM2VTS | |
|---|---|---|---|---|
| | Recognition accuracy | Running time | Recognition accuracy | Running time |
| PCA  | 0.77333 | 0.36752 | 0.71525 | 17.2797 |
| 2DPCA | 0.77333 | 0.29838 | 0.77797 | 2.4722 |

### 4.2. R$_1$-PCA and 2DR$_1$-PCA

The experiments result of R$_1$-PCA and 2DR$_1$-PCA is shown in Table 3. The two methods both iterate 120 times.

Table 3. Experiments result of R$_1$-PCA and 2DR$_1$-PCA

| Algorithms | ORL | | YALE | | XM2VTS | |
| --- | --- | --- | --- | --- | --- | --- |
| | Recognition accuracy | Running time | Recognition accuracy | Running time | Recognition accuracy | Running time |
| R$_1$-PCA | 0.88 | 914.2168 | 0.77333 | 411.0627 | 0.7161 | 1409.308 |
| 2DR$_1$-PCA | 0.905 | 403.9035 | 0.8 | 372.7688 | 0.7822 | 619.7837 |

Compared to PCA and 2DPCA, the running time of R$_1$-PCA and 2DR$_1$-PCA is much longer.

A start projection matrix $W^{(0)}$ is obtained by PCA (2DPCA) at the beginning of R$_1$-PCA (2DR$_1$-PCA). The final projection matrix $W$ is obtained by an iteration method starting with $W^{(0)}$. As a result of the iteration, the computational complexity is high.

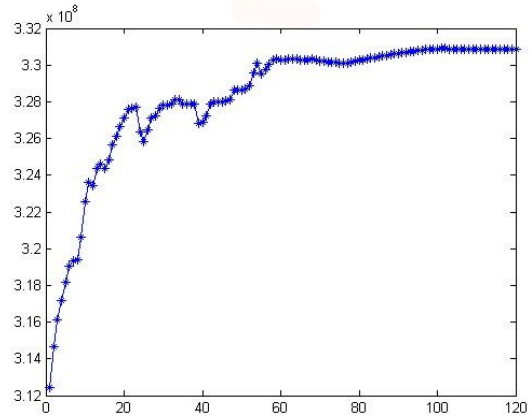

Fig. 1. The convergence illustration of iterating 120 times in R$_1$-PCA on the ORL database.

In R$_1$-PCA algorithm tested on the ORL database, the convergence process of iterating 120 times is shown in Fig. 1. After iterations at least 100 times the projection matrix $W$ is convergent. As a comparison, the corresponding illustration of 2DR$_1$-PCA is shown in Fig. 2.

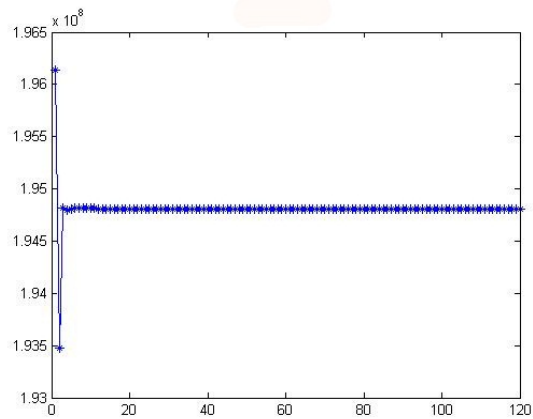

Fig. 2. The convergence illustration of iterating 120 times in 2DR$_1$-PCA on the ORL database.

From Fig. 2 we can see that after 20 iterations the projection matrix $W$ is convergent. The following 100 iterations are not needed. The computational time is wasted. In fact, a little number of iterations could get a convergent projection matrix. Fig. 3 shows the illustration of iterating 30 times.

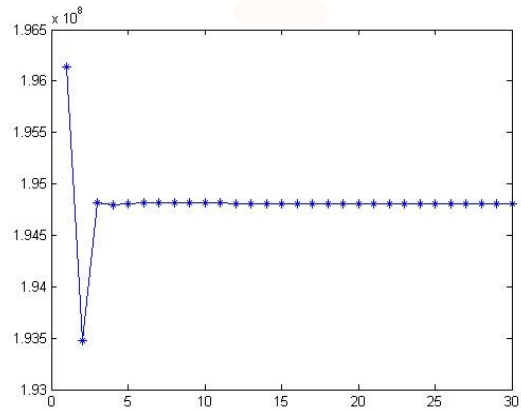

Fig. 3. The convergence illustration of iterating 30 times in 2DR$_1$-PCA on the ORL database.

From Fig. 3 we can see that the projection matrix $W$ is convergent when iterate 20 times. The little number of iterations would lead to a higher efficiency. The experimental result of iterating 120 times and 30 times in 2DR$_1$-PCA respectively is shown in Table 3.

Table 3. Experimental result of iterating 120 times and 30 times respectively in 2DR$_1$-PCA.

| Numbers of iterations | ORL | | YALE | | XM2VTS | |
|---|---|---|---|---|---|---|
| | Recognition accuracy | Running time | Recognition accuracy | Running time | Recognition accuracy | Running time |
| 120 | 0.905 | 403.904 | 0.8 | 372.769 | 0.7822 | 619.784 |
| 30 | 0.905 | 98.6516 | 0.8 | 90.0155 | 0.7822 | 162.6 |

From Table 3 we can see that in the different cases of iterating 120 times and 30 times, the recognition accuracy is the same, but the running time is quite different. In other words, the 2DR$_1$-PCA needs less number of iterations to obtain a convergent projection matrix. The efficiency of 2DR$_1$-PCA is much higher than that of R$_1$-PCA.

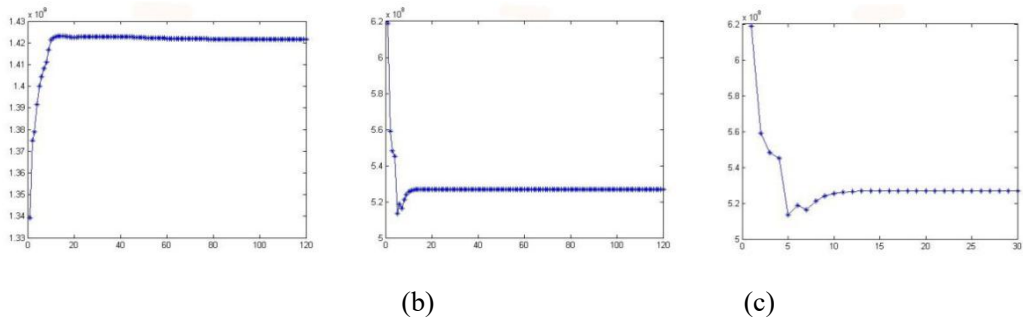

(a)　　　　　　　　　　　　(b)　　　　　　　　　　　　(c)

Fig. 4. The convergence illustration of R$_1$-PCA and 2DR$_1$-PCA tested on the Yale database. (a) R$_1$-PCA. (b) 2DR$_1$-PCA iterates 120 times. (c) 2DR$_1$-PCA iterates 30 times.

The convergence process of R$_1$-PCA and 2DR$_1$-PCA tested on the Yale database is shown in Fig. 4. The convergence speed of R$_1$-PCA is fast on the Yale database. The performance of R$_1$-PCA is close to that of 2DR$_1$-PCA.

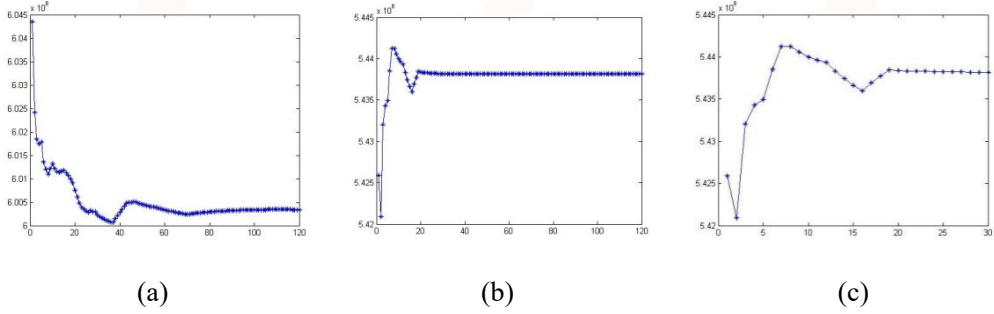

(a)                                (b)                              (c)

Fig. 5. The convergence illustration of $R_1$-PCA and $2DR_1$-PCA tested on the XM2VTS database. (a) $R_1$-PCA. (b) $2DR_1$-PCA iterates 120 times. (c) $2DR_1$-PCA iterates 30 times.

The comparison of convergence process in $R_1$-PCA and $2DR_1$-PCA on the XM2VTS database is shown in Fig. 5. The convergence speed of $2DR_1$-PCA is faster than that of $R_1$-PCA.

### 4.3. $L_1$-PCA and $2DL_1$-PCA

The experimental result of $L_1$-PCA and $2DL_1$-PCA are shown in Table 4.

Table 4. Experimental results of $L_1$-PCA and $2DL_1$-PCA.

| Algorithms | ORL | | YALE | | XM2VTS | |
|---|---|---|---|---|---|---|
| | Recognition accuracy | Running time | Recognition accuracy | Running time | Recognition accuracy | Running time |
| $L_1$-PCA | 0.885 | 15.9625 | 0.78667 | 3.3095 | 0.71695 | 83.5224 |
| $2DL_1$-PCA | 0.915 | 43.877 | 0.8 | 21.9788 | 0.74576 | 116.3024 |

The running time of $2DL_1$-PCA is longer than that of $L_1$-PCA. For the training samples $F_1, F_2, F_3, \cdots, F_i, \cdots, F_n$ where $F_i$ is a $m \times n'$ matrix, the training set used in $L_1$-PCA are $n$ vectors and used in $2DL_1$-PCA are $n' \times n$ vectors. The numbers of iterations in $2DL_1$-PCA are more than that in $L_1$-PCA. Thus the efficiency of $2DL_1$-PCA is lower. On the other hand, image matrices contain more information than image vectors. The recognition accuracy of $2DL_1$-PCA is higher than that of $L_1$-PCA.

### 4.4. KPCA and KECA

The contrast experimental result of KPCA and KECA is shown in Table 5.

Table 5. Experimental results of KPCA and KECA.

| Algorithms | ORL | | YALE | | XM2VTS | |
|---|---|---|---|---|---|---|
| | Recognition accuracy | Running time | Recognition accuracy | Running time | Recognition accuracy | Running time |
| KPCA | 0.925 | 5.8988 | 0.8533 | 2.5547 | 0.76441 | 105.6793 |
| KECA | 0.93 | 5.5516 | 0.8667 | 2.1784 | 0.79661 | 99.8656 |

Those principal axes contributing most to the Renyi entropy estimate clearly carry most of the information [6]. The projection matrix in KECA is formed based on the Renyi entropy contribution. Thus the recognition accuracy of KECA is higher than that of KPCA.

### 4.5. ECA and 2DECA

The experimental results of ECA and 2DECA are shown in Table 6.

Table 6. Experimental results of ECA and 2DECA.

| Algorithms | ORL | | YALE | | XM2VTS | |
|---|---|---|---|---|---|---|
| | Recognition accuracy | Running time | Recognition accuracy | Running time | Recognition accuracy | Running time |
| ECA | 0.91 | 4.5485 | 0.77333 | 2.2401 | 0.75085 | 25.1341 |
| 2DECA | 0.91 | 0.21218 | 0.77333 | 0.32116 | 0.75085 | 0.2864 |

Both ECA and 2DECA select eigenvectors based on the entropy contribution. The same recognition accuracy is obtained in the experiments. However, the efficiency is quite different. The running time of 2DECA is much shorter than that of ECA. The 2-D algorithm has performance advantages compared to its corresponding 1-D algorithm.

### 4.6. PCA and ECA

The experimental results of PCA and ECA are shown in Table 7.

Table 7. Experimental result of PCA and ECA.

| Algorithms | ORL | | YALE | | XM2VTS | |
|---|---|---|---|---|---|---|
| | Recognition accuracy | Running time | Recognition accuracy | Running time | Recognition accuracy | Running time |
| PCA | 0.9 | 1.375 | 0.77333 | 0.36752 | 0.71525 | 17.2797 |
| ECA | 0.91 | 4.5485 | 0.77333 | 2.2401 | 0.75085 | 25.1341 |

From Table 7 we can see that the recognition accuracy of ECA is higher than that of PCA because of forming projection matrix based on the Renyi entropy contribution in ECA. An additional step of calculating the entropy is contained in ECA, thus the running time of ECA is a little longer than that of PCA.

### 4.7. Comparisons Between All of The Methods

The experimental results of all of the mentioned methods are shown in Table 8. The numbers behind $R_1$-PCA and $2DR_1$-PCA denote the number of iterations.

Table 8. Experimental results of the mentioned methods.

| Algorithms | | ORL | | YALE | | XM2VTS | |
|---|---|---|---|---|---|---|---|
| | | Recognition accuracy | Running time | Recognition accuracy | Running time | Recognition accuracy | Running time |
| PCA | | 0.9 | 1.375 | 0.77333 | 0.36752 | 0.71525 | 17.2797 |
| 2DPCA | | 0.91 | 0.2913 | 0.77333 | 0.29838 | 0.77797 | 2.4722 |
| $R_1$-PCA | 120 | 0.88 | 914.2168 | 0.77333 | 411.0627 | 0.7161 | 1409.308 |
| $2DR_1$-PCA | 120 | 0.905 | 403.9035 | 0.8 | 372.769 | 0.7822 | 619.7837 |
| | 30 | 0.905 | 98.6516 | 0.8 | 90.0155 | 0.7822 | 162.6001 |
| $L_1$-PCA | | 0.885 | 15.9625 | 0.78667 | 3.3095 | 0.71695 | 83.5224 |
| $2DL_1$-PCA | | 0.915 | 43.877 | 0.8 | 21.9788 | 0.74576 | 116.3024 |
| KPCA | | 0.925 | 5.8988 | 0.8533 | 2.5547 | 0.76441 | 105.6793 |
| KECA | | 0.93 | 5.5516 | 0.8667 | 2.1784 | 0.79661 | 99.8656 |
| ECA | | 0.91 | 4.5485 | 0.77333 | 2.2401 | 0.75085 | 25.341 |
| 2DECA | | 0.91 | 0.21218 | 0.77333 | 0.32116 | 0.75085 | 0.2864 |

Because it needs to iterate several times, the efficiency of $R_1$-PCA, $2DR_1$-PCA, $L_1$-PCA and $2DL_1$-PCA is lower than that of the traditional PCA algorithm. Nevertheless, their recognition accuracy is higher than that of PCA.

The samples are linear irreducible in the input space. In KPCA algorithm, the samples are projected into a high dimensional kernel space which is linear separable. Thus the recognition accuracy is improved by KPCA.

In KECA algorithm, the eigenvectors are selected based on the Renyi entropy contribution. These eigenvectors carry most of the information. Thus a higher recognition accuray is obtained in the KECA algorithm. And so do ECA and 2DECA.

## 5. CONCLUSIONS

After studying the PCA, 2DPCA, $R_1$-PCA, $L_1$-PCA and KECA algorithms, in this paper we propose $2DR_1$-PCA and $2DL_1$-PCA algorithms. Inspired by KECA, we use Renyi entropy in PCA and 2DPCA to obtain the projection matrix. Then the ECA and 2DECA algorithms are presented.

We analyze the performance between 1-D and their corresponding 2-D algorithm based on experiments. Generally speaking, 2-D algorithms have a higher recognition accuracy than that of the corresponding 1-D algorithms. The efficiency of 2-D algorithms is higher than that of their corresponding 1-D algorithms. An exception is $2DL_1$-PCA. Because of more iterations required according to Eq. (26), its efficiency is lower than that of $L_1$-PCA.

## ACKNOWLEDGEMENTS

The paper is supported by the National Natural Science Foundation of China (Grant No.61672265 and U1836218) and the 111 Project of Ministry of Education of China (Grant No. B12018).